\documentclass[10pt,twocolumn,letterpaper]{article}

\usepackage{wacv}              

\usepackage{booktabs}
\usepackage{multirow}
\usepackage{graphicx}
\usepackage{array}
\usepackage{anyfontsize}
\usepackage{makecell}
\usepackage{amsmath}
\usepackage{enumitem}
\usepackage{listings}
\usepackage{ulem}
\newcolumntype{H}{>{\setbox0=\hbox\bgroup}c<{\egroup}@{}}

\usepackage[pagebackref,breaklinks,colorlinks]{hyperref}

\usepackage[capitalize]{cleveref}
\crefname{section}{Sec.}{Secs.}
\Crefname{section}{Section}{Sections}
\Crefname{table}{Table}{Tables}
\crefname{table}{Tab.}{Tabs.}

\def\wacvPaperID{1128} 
\def\confName{WACV}
\def\confYear{2025}

\begin{document}

\title{SCOT: Self-Supervised Contrastive Pretraining For Zero-Shot Compositional Retrieval}

\author{Bhavin Jawade$^{1,2*}$ , Jo\~{a}o V. B. Soares$^{1}$, Kapil Thadani$^{1}$, Deen Dayal Mohan$^{1}$,\\
Amir Erfan Eshratifar$^{1}$, Benjamin Culpepper$^{1}$, Paloma de Juan$^{1}$,\\
Srirangaraj Setlur$^{2}$, Venu Govindaraju$^{2}$\\
$^{1}$Yahoo Research, $^{2}$University at Buffalo, SUNY\\
{\tt\small bhavinja@buffalo.edu, jvbsoares@yahooinc.com, thadani@yahooinc.com} \\
{\tt\small deendayal.mohan@yahooinc.com, erfan.eshratifar@yahooinc.com, jackcul@yahooinc.com} \\
{\tt\small pdjuan@yahooinc.com, setlur@buffalo.edu, govind@buffalo.edu}
}

\maketitle

\begin{abstract}
Compositional image retrieval (CIR) is a multimodal learning task where a model combines a query image with a user-provided text modification to retrieve a target image. CIR finds applications in a variety of domains including product retrieval (e-commerce) and web search.
Existing methods primarily focus on fully-supervised learning, wherein models are trained on datasets of labeled triplets such as FashionIQ and CIRR. This poses two significant challenges: (i) curating such triplet datasets is labor intensive; and (ii) models lack generalization to unseen objects and domains. In this work, we propose SCOT (Self-supervised COmpositional Training), a novel zero-shot compositional pretraining strategy that combines existing large image-text pair datasets with the generative capabilities of large language models to contrastively train an embedding composition network. Specifically, we show that the text embedding from a large-scale contrastively-pretrained vision-language model can be utilized as proxy target supervision during compositional pretraining, replacing the target image embedding. 
In zero-shot settings, this strategy surpasses SOTA zero-shot compositional retrieval methods as well as many fully-supervised methods on standard benchmarks such as FashionIQ and CIRR. Our code and models are available at \url{https://github.com/yahoo/SCOT}. 
\end{abstract}

\vspace{-0.5em}
\section{Introduction}
\footnotetext[1]{*Work done during research internship at Yahoo Research.}








\label{sec:intro}
The field of image retrieval is advancing rapidly, with growing interest in multimodal queries that incorporate both images and text. Compositional Image Retrieval (CIR)
is a recently proposed task that aims at retrieving images using a query composed of both an image and text~\cite{TIRG,guo2018dialog}. 
The query or reference image defines some initial desired elements, while the text describes the relative modification 
that a user would like to see in the retrieved images.
CIR provides users with a versatile way to communicate their intent through iterative query refinement, which is potentially valuable in a broad range of real-world tasks such as product retrieval in e-commerce and fine-grained web search. 

\begin{figure}[t]
\centering
\includegraphics[scale=0.11]{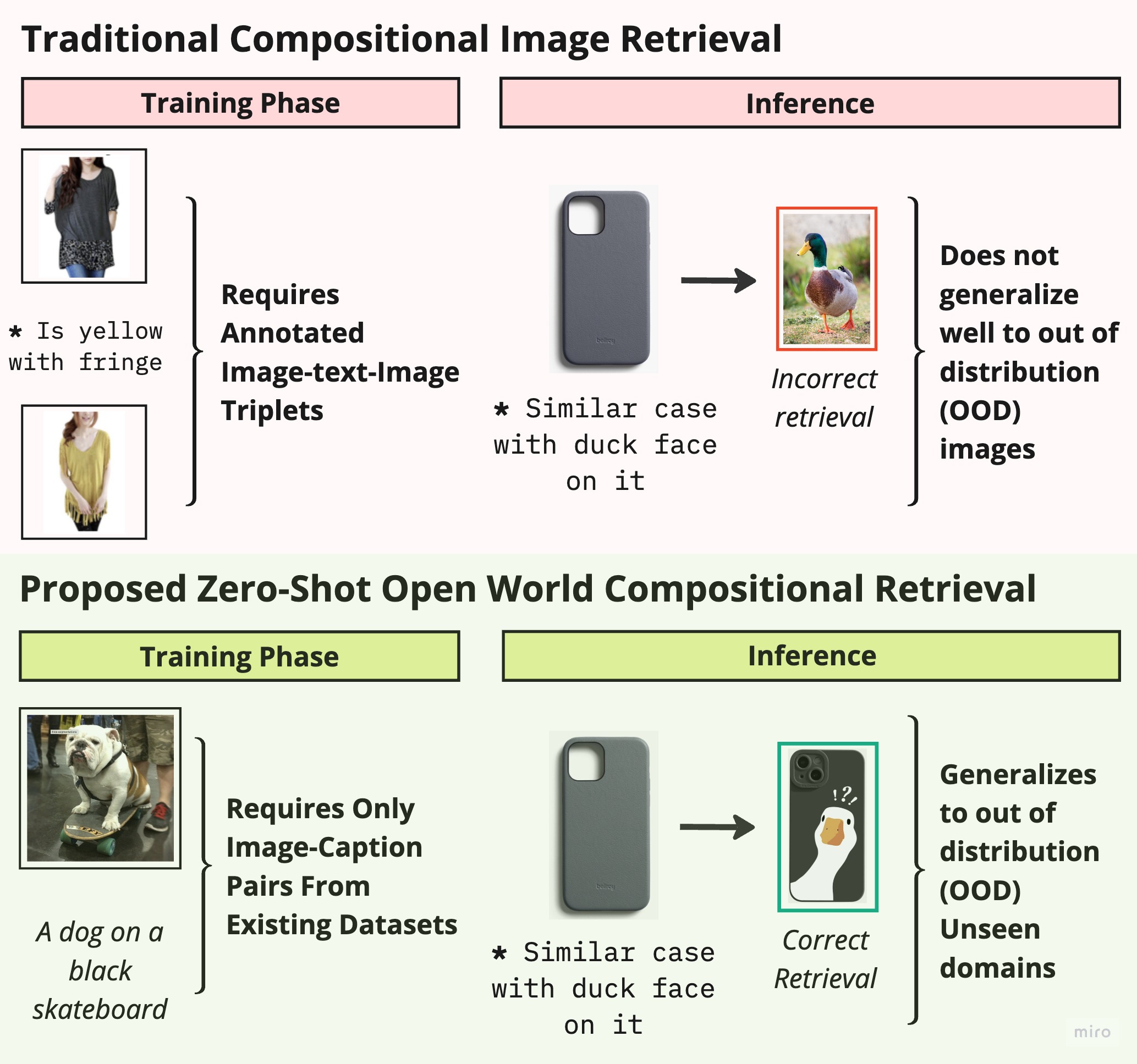}
\caption{Compositional image retrieval methods typically require domain-specific image-text-image triplets for training and cannot generalize to unseen domains. In contrast, SCOT uses existing large noisy captioned image datasets for compositional training and demonstrates zero-shot generalizability to new domains.}
\label{fig:conceptdiagram}
\vspace{-1.0em}
\end{figure}

CIR can be framed as a multimodal fused representation learning task in which the goal is to train an effective feature fusion network.
This sits in contrast to other well-studied vision-language tasks such as image-text matching, image captioning, and visual question answering, as CIR uniquely learns a representation to jointly capture visual cues and text descriptors that match the target image of interest. Most CIR methods \cite{TIRG, fashionVLP, dcnet, maaf} 
are trained in a fully-supervised manner using curated human-annotated datasets of triplets, with each triplet consisting of a reference image, a user-provided modification text, and a target image.

Current supervised CIR approaches do not generalize well to unseen domains or zero-shot scenarios, as illustrated in Fig~\ref{fig:conceptdiagram}. They are dependent on the availability of large datasets of image-text-image triplets, which are typically domain-specific and have limited applicability to open-world settings. Manual labeling for new triplet datasets is also labor-intensive. To overcome these challenges, a recent line of work explores zero-shot CIR using textual inversion~\cite{PALAVRA,Saito_2023_CVPR,SEARLE,tang2023context}, e.g., using image-text pairs to learn to map images into text token embeddings. An image-derived token embedding---which can be thought of as corresponding to a {\it pseudo-token})---can then be combined with text token embeddings from the modification text and encoded as text to produce a composite embedding for retrieval. These approaches do not require annotated image-text-image triplets and can adapt to new domains thanks to the generalizability of contrastively-pretrained image-text encoders.


In this work, we propose a novel pretraining strategy for zero-shot CIR (ZS-CIR) which we name SCOT (Self-supervised COmpositional Training). This approach does not require human-annotated triplets and demonstrates open-world generalizability by using captioned images from large and varied datasets. We specifically exploit the proximity of visual and textual representations of the same concept in the embedding space of large-scale contrastively-pretrained vision-language models, which enables the use of target text embeddings instead of target image embeddings
for supervision. Given an image and its caption, we first generate a training example by feeding the caption into a large language model (LLM) and prompting it to output a creative modification text and a corresponding modified caption. A CIR model is then trained by using the reference image and the generated modification text as input, with the generated modified caption as the target.

SCOT models are trained to compose reference images with modification texts by optimizing a contrastive image retrieval loss. This differs from inversion-based techniques, which do not directly train a composition model but rely on the composition capabilities of existing frozen pretrained image-text encoders. SCOT pretraining is agnostic to the choice of composition model, which can include unfrozen encoders~\cite{transagg} and early image-text fusion~\cite{caselasco}. 
Comprehensive experiments show that SCOT surpasses current ZS-CIR techniques and nears fully-supervised performance on FashionIQ~\cite{fashionIQ} and CIRR~\cite{CIRR} without domain-specific training.
%
%
The key contributions of this work are:
\vspace{-0.5em}
\begin{enumerate}[noitemsep]
    \item We introduce a novel compositional pretraining strategy that requires only image-text pairs, using LLMs to create image-text-text triplets and pretrained vision-language models to encode both images and text. 
    \item We demonstrate zero-shot generalizability on domain-specific (FashionIQ~\cite{fashionIQ}) and open-world (CIRR~\cite{CIRR}) compositional retrieval datasets, showing that SCOT outperforms existing zero-shot approaches. 
    \item Through quantitative and qualitative experiments, we evaluate the impact of various parameters such as training dataset size, sample distribution, backbone and supervision type on zero-shot generalizability. 
\end{enumerate}

\section{Related Work}
\label{sec:related}

\textbf{Compositional Image Retrieval (CIR):} Numerous methods have been proposed to learn composite representations of visual and text features for retrieval. Most research lies within the supervised setting~\cite{dcnet, CIRR, BLIP4Cir, combinerarch, TIRG, fashionVLP, maaf, delmas2022artemis, caselasco, tian2023fashion, barbany2024leveraging, jang2024visual}, with earlier work
relying on fashion datasets containing human-annotated triplets~\cite{fashionIQ,han2017automaticfashion200k}.
The DCNet approach~\cite{dcnet} jointly trains feature extractors with a composition and correction network on FashionIQ \cite{fashionIQ}. CoSMo\cite{cosmo} uses content and style modulator networks to combine the image and text representations. FashionVLP~\cite{fashionVLP} is a recently-proposed multimodal Transformer trained with a variety of fashion image inputs including crops, landmarks and ROIs. The need to go beyond fashion products and motivate research in open-world interactive retrieval led to the creation of open-domain annotated datasets: CIRR~\cite{CIRR} (using images from NLVR2~\cite{nlvr2}), CIRCO~\cite{SEARLE}, and LaSCo~\cite{caselasco} (the latter two using images from MS-COCO~\cite{mscoco}). Despite this progress, the zero-shot generalizability of traditional fully-supervised models has been limited.

\begin{figure*}[h]
\centering
\includegraphics[scale=0.49]{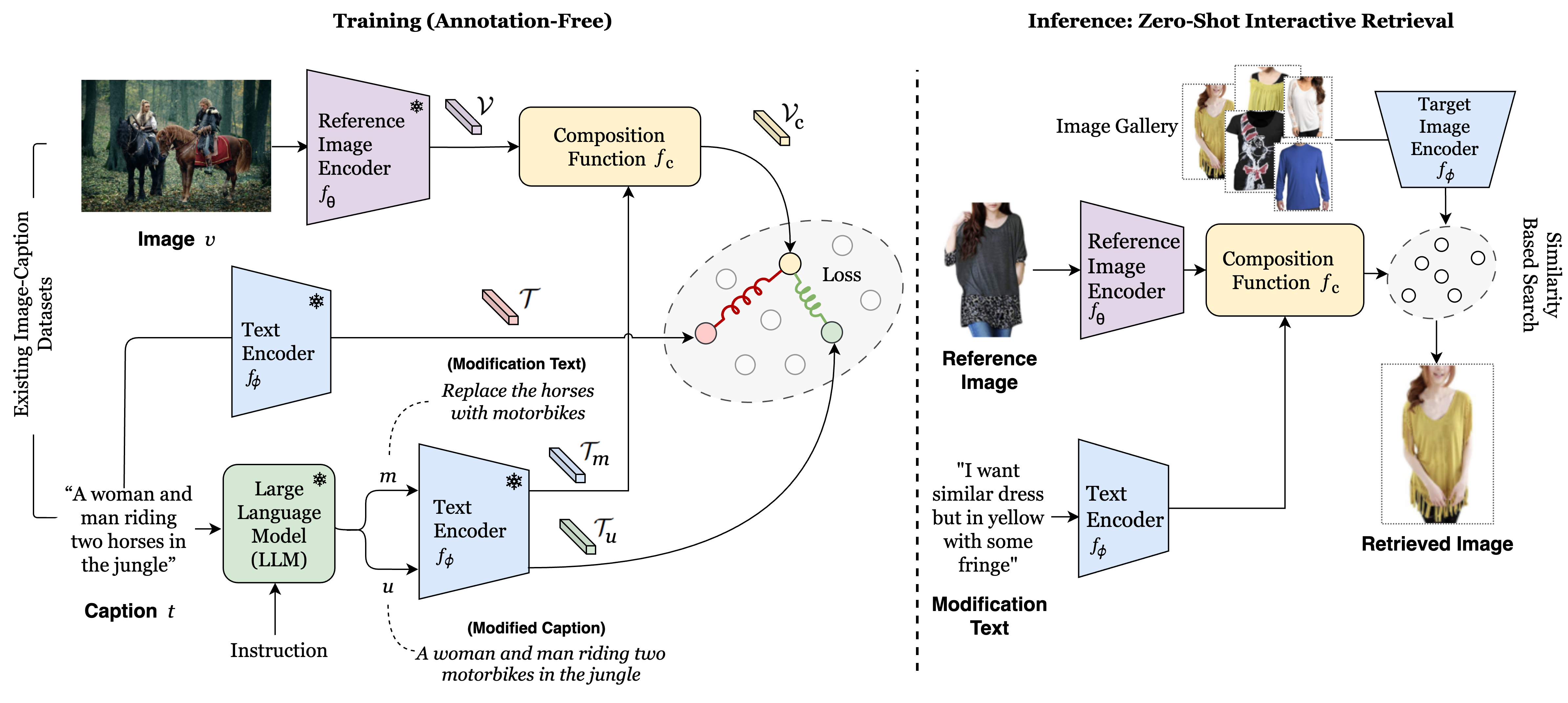}
\caption{\textbf{SCOT pretraining and inference.} \textbf{Left:} The composition function $f_c$ is trained using existing image-caption datasets, a frozen image-text encoder (such as CLIP), and a frozen large language model (LLM). The LLM generates the modification text $m$ and a modified caption $u$. The reference image embedding $\mathcal{V}$ and the modification text embedding $\mathcal{T}_m$ are passed to $f_c$ to get the composed embedding $\mathcal{V}_c$. We optimize the parameters of $f_c$ to draw $\mathcal{V}_c$ towards the modified caption $\mathcal{T}_u$ and away from the original caption $\mathcal{T}$. The full loss also pushes $\mathcal{V}_c$ away from the embeddings of other (non-matching) modified captions within each batch (not illustrated here). \textbf{Right:} During inference, we compute the similarity between the composed embedding and the embeddings of gallery images to retrieve the target image.}
\label{fig:architecture}
\vspace{-0.5em}
\end{figure*}

\noindent\textbf{Zero-Shot Compositional Retrieval (ZS-CIR):} To overcome these limitations, recent work \cite{Saito_2023_CVPR, PALAVRA, SEARLE, trainingfree, transagg, chen2023pretrain, tang2023context, gu2023compodiff} has developed zero-shot annotation-free strategies for CIR.
%
One line of work~\cite{PALAVRA,Saito_2023_CVPR,SEARLE,tang2023context} adopts textual inversion, which had previously found success in the text-to-image generation literature~\cite{gal2022textual}.
Recently, Saito et al.~\cite{Saito_2023_CVPR} proposed Pic2Word wherein an MLP is trained to map a picture to a pseudo-token, which the text encoder can then combine with the modification text to produce a composite embedding.
Baldrati et al.~\cite{SEARLE} present SEARLE, which involves a two-stage process for training a textual inversion network. The first stage runs Optimization-based Textual Inversion (OTI) 
with CLIP~\cite{CLIPopenai} image and text encoders to find a text token embedding that corresponds to a given image encoding. In the second stage, those token embeddings are used as targets to learn a textual inversion network. 
Note that textual inversion approaches focus on learning how to invert the image into token embeddings, while taking advantage of the existing composition capabilities of pretrained text encoders. In contrast, our approach directly optimizes a contrastive loss by training with triplets that closely mimic those of the CIR task. It can thus use any choice of composition model (including unfrozen encoders), and can be easily fine-tuned further with domain-specific data.
A variety of other ZS-CIR approaches have been recently proposed. Gu et al.~\cite{gu2023compodiff} train a denoising Transformer for image-text composition on 18M synthetic images along with 2B captioned images from LAION~\cite{schuhmann2022laion}. Karthik et al.~\cite{trainingfree} introduce CIReVL, a training-free approach that involves captioning the reference image, modifying the caption using an LLM and retrieving the target image using the modified caption. Chen and Lai~\cite{chen2023pretrain} propose masking-augmented contrastive pretraining for visual and textual encoders to recover masked visual information through text prompts.
Jang et al.~\cite{jang2024visual} train a model to generate the modification text given a pair of images. The model is sued for generating synthetic training data, resulting in a semi-supervised approach.
In concurrent work, Liu et al.~\cite{transagg} propose an approach for automatic construction of image-text-image training triplets. They source captioned images from the LAION-COCO dataset~\cite{laioncoco} and use either text templates or LLMs to generate modification texts and corresponding modified captions. Modified captions are then used to retrieve images to serve as supervision targets.
The authors note that this approach of retrieving supervision target images from a corpus can be problematic due to the eventual absence of suitable images and/or retrieval errors~\cite{transagg}.
In Section~\ref{sec:discussion}, we show example triplets illustrating these issues and
present a controlled experiment demonstrating that SCOT's use of semantically-relevant text targets significantly outperforms the use of retrieved image targets.

\section{Method}
\label{sec:method}


This section describes SCOT, a ZS-CIR technique requiring only captioned image datasets.
The approach is outlined in Fig.~\ref{fig:architecture}.
We review contrastively pretrained image-text encoders in Section~\ref{sec:largescale}. 
Sections~\ref{sec:selfsupervised}, \ref{sec:objective} and~\ref{sec:inference} detail our pretraining strategy, loss function, and inference.




\subsection{Large-Scale Contrastive Pretraining}
\label{sec:largescale}


Following previous work, we use image and text representations from large-scale contrastively-pretrained models: CLIP~\cite{CLIPopenai}, BLIP~\cite{li2022blip1} and BLIP-2~\cite{li2023blip2}.
CLIP (Contrastive Language-Image Pretraining)~\cite{CLIPopenai} aims to jointly learn visual and textual representations that are semantically aligned. For a given image-caption pair $(v_i, t_i)$, let $\mathcal{V}_i = f_\theta(v_i)$ denote the normalized image embedding from image encoder $f_\theta$ and $\mathcal{T}_i = f_\phi(t_i)$ denote the normalized text embedding from text encoder $f_\phi$. CLIP contrastively enforces high similarity between positive pairs $\left(\mathcal{V}_i, \mathcal{T}_i\right)$ and low similarity between negative pairs $\left(\mathcal{V}_i, \mathcal{T}_j\right), \  \forall \ i \neq j$. This is implemented via a symmetric cross-entropy loss over the similarity scores of image and text embeddings $\mathcal{V}_i$ and $\mathcal{T}_j$. The image-to-text part of the loss is defined as:
\begin{align}
\mathcal{L}_{\text{i2t}} &= -\frac{1}{N}\sum_{i=1}^{N} \log \frac{e^{\langle \mathcal{V}_i, \mathcal{T}_i \rangle / \kappa}}{\sum_{j=1}^{N} e^{\langle \mathcal{V}_i, \mathcal{T}_j \rangle / \kappa}}
\end{align}
where \(\langle \cdot, \cdot \rangle\) is the dot product, \(N\) the batch size and \(\kappa\) the temperature parameter.

\begin{figure}
\centering
\includegraphics[scale=0.12]{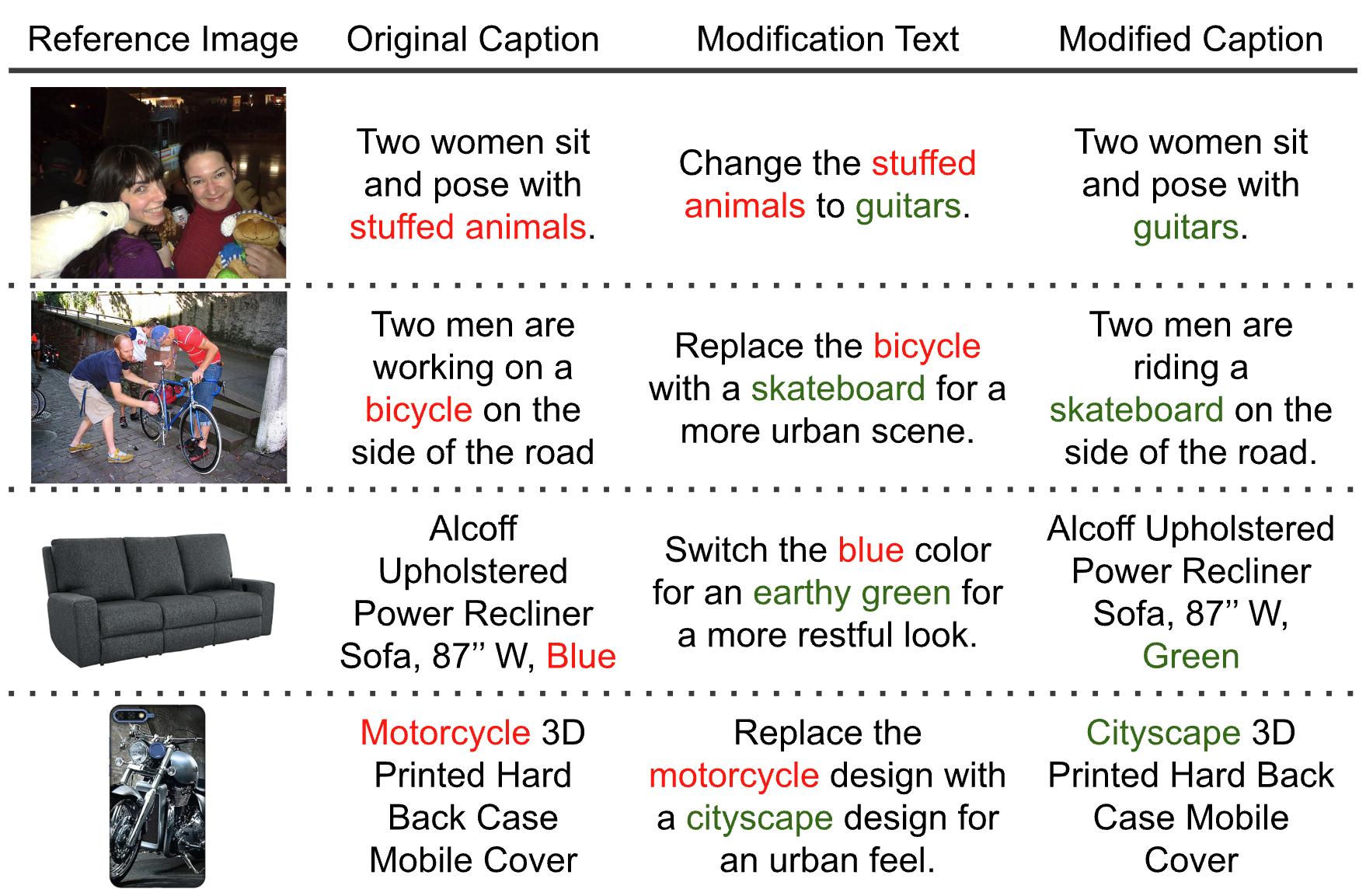}
\caption{\textbf{LLM-generated text triplet samples,} showing appropriate modifications over different image domains.}
\label{fig:llm_samples}
\vspace{-1.0em}
\end{figure}

BLIP~\cite{li2022blip1} and BLIP-2~\cite{li2023blip2} are other pretraining approaches that 
demonstrate strong performance on benchmarks for image-text retrieval.
BLIP-2 employs a lightweight trainable Querying Transformer (Q-Former) module whose image features come from a frozen pretrained CLIP encoder. The initial training stage performs representation learning by jointly optimizing three objectives that include an image-text contrastive loss as in CLIP.


\subsection{Self-Supervised Compositional Pretraining}
\label{sec:selfsupervised}

Our approach is primarily motivated by the fact that contrastively-pretrained models are able to align related visual and textual representations in the embedding space. This enables us to use the aligned textual representation as a proxy for an image representation, thereby eliminating the need for a target image during training. For inference, we can search across gallery images by encoding them using the 
the contrastively-paired visual encoder.

The goal of this method is to train a composition operation $f_c$ to 
combine the representations from a user-provided image and modification text.
We rely on contrastively-paired image and text encoders, denoted respectively as $f_\theta$ and $f_\phi$. Given a captioned image dataset $D = \{(v_i, t_i)\}_{i=1}^{M}$, we compute the image embeddings $\mathcal{V}^i = f_\theta(v_i)$
and corresponding caption embeddings $\mathcal{T}^i = f_\phi(t_i)$. 

We prompt a large language model (LLM) to generate a modification text $m_i$ given an original caption $t_i$. The modification text will be used as one of the inputs to the composition function $f_c$ during training. To provide the supervision signal for the predicted composed representation, 
we use the same LLM to generate a modified caption $u_i$, which should be similar to $t_i$ but with modification $m_i$ applied. Fig.~\ref{fig:llm_samples} contains samples of LLM-generated triplets. Next, we compute the embeddings for $m_i$ and $u_i$ using $f_\phi$. 
\begin{align}
m_i,\; u_i &\leftarrow \mbox{LLM}(t_i) \\
\mathcal{T}_m^i,\; \mathcal{T}_u^i &= f_\phi(m_i),\; f_\phi(u_i)
\end{align}
We pass the modification text representation $\mathcal{T}_m^i$ and the image representation $\mathcal{V}^i$ through the learnable composition function~$f_c$ to obtain the composed image representation $\mathcal{V}_c^i = f_c\left(\mathcal{V}^i, \mathcal{T}_m^i\right)$.
We use the recently proposed Combiner network~\cite{combinerarch} to implement $f_c$. Briefly, it performs a learnable weighted fusion of the image and text embeddings. We encourage readers to refer to~\cite{combinerarch} or the supplementary material of this paper for details on the Combiner network. 

\subsection{Training Objective}

\label{sec:objective}

We minimize a modified contrastive loss in order to pull the predicted composed embedding $\mathcal{V}_c^i$ towards the generated target text embedding $\mathcal{T}_u^i$ for an input sample ($v_i$, $t_i$) while pushing it away from target text embeddings $\mathcal{T}_u^j,\ \forall j \neq i$ from other examples  within its batch. Let \(S(x, y)\) denote the cosine similarity between vectors \(x\) and \(y\), i.e., $S(x, y) = \frac{x \cdot y}{\|x\|_2 \|y\|_2}$. We define:
\vspace{-0.5em}
\begin{align}
\mathcal{L}_{\text{pos}} &= - \log \sum_{i=1}^N e^{S\left(\mathcal{V}_c^i, \mathcal{T}_u^i\right)}
\label{ref:pos_eq}\\
\mathcal{L}_{\text{neg}} & = \log \sum_{i, j}^N e^{ S\left(\mathcal{V}_c^i, \mathcal{T}_u^j\right) \cdot (1 - \delta_{ij})}
\label{ref:neg_eq}
\end{align}
\noindent where \(\delta_{ij}\) is the Kronecker delta function, which is 1 when \(i = j\) and 0 otherwise.

Previous work in traditional cross-modal retrieval \cite{vseplusplus, wei2020universal} has demonstrated the effectiveness of hard-negative mining. To improve the robustness of our embeddings, we follow \cite{wang2014learning, vseplusplus} and adopt a \textit{margin-based} hard-negative mining strategy. Let $\lambda$ be a fixed scalar margin, then we define:
\begin{align}
S_\lambda(x, y) &= S(x, y) \cdot \mathrm{\Theta}\left(S(x, y) > \lambda\right)
\end{align}
where $\mathrm{\Theta}$ is the Heaviside step function, which is 1 if the condition inside is true and 0 otherwise. This is used in Eq.~\ref{ref:neg_eq} resulting in an updated negative loss.
\vspace{-0.5em}
\begin{align}
\mathcal{L}_{\text{neg}}' & = \log \sum_{i, j}^N e^{ S_\lambda\left(\mathcal{V}_c^i, \mathcal{T}_u^j\right) \cdot (1 - \delta_{ij})}
\label{ref:hardneg_eq}
\end{align}
For stronger supervision, we also include the original unmodified caption embeddings $\mathcal{T}^j \ \forall j \leq N$ as hard negatives for $\mathcal{V}_c^i$. This moves the composed representation away from the original caption and closer to the desired modified caption, ensuring it does not retain features from the original sample that are absent in the target caption. We use all the original captions in a batch as negatives for that batch, resulting in the following combined loss for negatives.
\begin{align}
\mathcal{L}_{\text{neg}}'' = \mathcal{L}_{\text{neg}}' + \log \sum_{i, j}^N e^{S_\lambda(\mathcal{V}_c^i, \mathcal{T}^j)}
\label{ref:combinedneg_eq}
\end{align}
%
%
Using Eqs. (\ref{ref:pos_eq}, \ref{ref:combinedneg_eq}) we minimize the following final loss with respect to the parameters of the composition function $f_c$:
\begin{align}
\mathcal{L} &= \alpha_{\text{pos}} \cdot \mathcal{L}_{\text{pos}} + \alpha_{\text{neg}} \cdot \mathcal{L}_{\text{neg}}''
\end{align}
where \(\alpha_{\text{pos}}\) and \(\alpha_{\text{neg}}\) are positive and negative scaling factors respectively. In summary, the composition function is trained to apply the LLM-generated modification text to the reference image such that the resulting composed representation lies close to the embedding of the modified caption.

\subsection{Inference}
\label{sec:inference}

As shown in Fig.~\ref{fig:architecture}, all gallery images for retrieval are encoded with the image encoder $f_\theta$. During inference, we combine the embeddings of the reference image and user-provided modification text using the learned composition function $f_c$, as in training. This composite representation is used to retrieve the most similar gallery images by computing cosine similarity with their image embeddings.

\section{Experiments}
\label{sec:experiments}
We now turn to quantitative and qualitative evaluations of SCOT for ZS-CIR. Additional results are in the appendix.

\subsection{Datasets}
\noindent We train on three datasets of captioned images: MS-COCO~\cite{mscoco} (189K pairs), Flickr30K~\cite{flickr30k} (45K pairs), and ABO~\cite{ABO} (58K pairs), totaling 290K image-text pairs. Following previous works \cite{SEARLE, transagg, Saito_2023_CVPR}, we assess zero-shot capabilities on FashionIQ~\cite{fashionIQ} and CIRR~\cite{CIRR}, two compositional retrieval datasets with annotated triplets. Here, FashionIQ assess zero-shot generalizability in the fashion domain and CIRR on open-world retrieval setting.

\subsection{Implementation Details}
\noindent \textbf{Encoders.} Unless otherwise stated, we use  
\ BLIP-2\footnote{We use BLIP-2 with EVA-CLIP ViT-G/14 backbone from \href{https://github.com/salesforce/LAVIS}{LAVIS}.} as a frozen\footnote{While encoders can also be finetuned with SCOT, we keep them frozen to compare fairly with prior work, most of which uses frozen encoders.} image and text encoder.

\noindent \textbf{Textual triplet generation.}
To generate modification texts $m$ and modified captions $u$, we use the instruction-tuned Falcon-7B LLM~\cite{almazrouei2023falcon}. As directly prompting this model produces noisy and inconsistent generations on our task, we generate 4K text triplets from the better-performing GPT-4 model~\cite{gpt4} and use them for LoRA fine-tuning~\cite{lora} of 4-bit quantized Falcon-7B \cite{almazrouei2023falcon}. Finally, we generate a dataset of over 290K text triplets using the finetuned Falcon-7B, which can be reused in subsequent training runs. 
SCOT is not reliant on any specific LLM, so newer or stronger models can also be used to refine and expand the triplet dataset.

\noindent \textbf{Other training details.} We train with AdamW~\cite{adamw}, batch size 1024 and learning rate 1x10$^{-4}$. In the loss (Sec.~\ref{sec:objective}), we set positive scaling factor $\alpha_{\text{pos}}=10$ and negative scaling factor $\alpha_{\text{neg}} = 0.1$, and margin $\lambda = 0.2$. For the Combiner, we use the same hyperparameters as the original work~\cite{combinerarch}. Training and inference uses 2 NVIDIA A100 GPUs. 

\begin{table*}[t]
\centering
\footnotesize
\begin{tabular}{@{}lllccccccccH@{}}

\toprule
\multirow{2}{*}{\textbf{}} & \multirow{2}{*}{\textbf{Backbone}} & \multirow{2}{*}{\textbf{Method}} & \multicolumn{2}{c}{\textbf{Average}} & \multicolumn{2}{c}{\textbf{Dress}} & \multicolumn{2}{c}{\textbf{Shirt}} & \multicolumn{2}{c}{\textbf{Top/Tee}} & \multirow{2}{*}{\textbf{FIQ}} \\
\cmidrule(lr){4-5} \cmidrule(lr){6-7} \cmidrule(lr){8-9} \cmidrule(lr){10-11}
& & & \textbf{R@10} & \textbf{R@50} & \textbf{R@10} & \textbf{R@50} & \textbf{R@10} & \textbf{R@50} & \textbf{R@10} & \textbf{R@50} & \\
\midrule
\parbox[t]{15pt}{\multirow{5}{*}{\rotatebox[origin=c]{90}{Superv.}}}
& Multi & MAAF \cite{maaf} & 24.3\phantom{0} & 48.8\phantom{0} & 23.8\phantom{0} & 48.6\phantom{0} & 21.3\phantom{0} & 44.2\phantom{0} & 27.9\phantom{0} & 53.6\phantom{0}\\ 
& Multi & DCNet \cite{dcnet} & 30.44 & 58.29 & 28.95 & 56.07 & 23.95 & 47.30 & 30.44 & 58.29 & 44.37\\ 
& Multi & FashionVLP \cite{fashionVLP} & 34.27 & 62.51 & 32.42 & 60.29 & 31.89 & 58.44 & 38.51 & 68.79 & 48.39\\
& CLIP L/14 & CLIP4CIR \cite{combinerarch} & 38.32 & 61.74 & 33.81 & 59.40 & 39.99 & 60.45 & 41.41 & 65.37 & 50.03\\
& BLIP & BLIP4CIR \cite{BLIP4Cir} & 43.49 & 67.31 & 42.09 & 67.33 & 41.76 & 64.28 & 46.61 & 70.32 & 55.40\\
\midrule

\parbox[t]{15pt}{\multirow{23}{*}{\rotatebox[origin=c]{90}{Zero-Shot}}}
& \multirow{7}{*}{CLIP B/32} & Image-Only & \phantom{0}5.88 & 13.19 &  \phantom{0}6.96 & 14.08 & \phantom{0}4.46 & 11.89 & \phantom{0}6.22 & 13.61 \\ 
& & Text-Only & 18.41 & 36.28 & 14.92 & 33.81 & 19.77 & 34.69 & 20.55 & 40.33 \\
& & Image+Text & 13.36 & 27.51 &  12.44 & 28.55 & 12.61 & 24.82 & 15.04 & 29.16  \\
& & PALAVRA \cite{PALAVRA} & 19.76 & 37.25 & 17.25 & 35.94 & 21.49 & 37.05 & 20.55 & 38.76 & 28.50 \\
& & SEARLE \cite{SEARLE} & 22.89 & 42.53 & 18.54 & 39.51 & 24.44 & 41.61 & 18.54 & 39.51 \\
& & TransAgg \cite{transagg} & \underline{23.91} & \textbf{44.68} & \underline{19.44} & \textbf{42.04} & \underline{25.37} & \underline{42.69} & \underline{26.93} & \textbf{49.31} \\
& & SCOT (Ours) & \textbf{24.14} & \underline{43.44} & \textbf{19.73} & \underline{41.24} & \textbf{25.51} & \textbf{42.93} & \textbf{27.18} & \underline{46.14} \\

\cmidrule(lr){2-11}

& \multirow{7}{*}{CLIP L/14} & Image-Only & \phantom{0}7.97 & 17.43 & \phantom{0}5.25 & 13.63 & 10.54 & 20.65 & \phantom{0}8.10 & 18.01  \\ 
& & Text-Only & 19.01 & 35.26 & 15.22 & 33.01 & 19.82 & 33.31 & 21.87 & 39.46 \\
& & Image+Text & 18.12 & 33.17 & 14.27 & 31.33 & 19.13 & 32.28 & 20.95 & 35.90 \\
& & Pic2Word \cite{Saito_2023_CVPR} & 24.7\phantom{0} & 43.7\phantom{0} & 20.0\phantom{0} & 40.2\phantom{0} & 26.2\phantom{0} & 43.6\phantom{0} & 27.9\phantom{0} & 47.4\phantom{0} & 34.2\phantom{0} \\
& & SEARLE-XL \cite{SEARLE} & 25.56 & 46.23 & 20.48 & 43.13 & 26.89 & 45.58 & 29.32 & 49.97 \\
& & TransAgg \cite{transagg} & \textbf{28.57} & \textbf{48.29} & \textbf{23.85} & \underline{44.57} & \textbf{29.54} & \textbf{47.79} & \textbf{32.33} & \textbf{52.52}  \\
& & SCOT (Ours) & \underline{28.27} & \underline{47.44} & \underline{23.69} & \textbf{45.06} & \underline{29.09} & \underline{47.01} & \underline{32.02} & \underline{50.33}  \\

\cmidrule(lr){2-11}

& \multirow{5}{*}{BLIP} & Image-Only & \phantom{0}6.65 & 15.40 & \phantom{0}5.05 & 12.19 & \phantom{0}7.55 & 17.76 & \phantom{0}7.34 & 16.26 \\
& & Text-Only & 24.01 & 42.73 & 20.03 & 39.96 & 24.63 & 41.02 & 27.38 & 47.22  \\
& & Image+Text & \phantom{0}8.06 & 18.16 & \phantom{0}6.14 & 19.78 & \phantom{0}9.37 & 19.87 & \phantom{0}8.66 & 19.78  \\
& & TransAgg \cite{transagg} & \underline{26.95} & \underline{46.10} & \underline{21.67} & \underline{41.89} & \underline{28.07} & \underline{45.63} & \underline{31.11} & \underline{50.79}  \\
& & SCOT (Ours) & \textbf{30.68} & \textbf{51.33} & \textbf{26.42} & \textbf{49.23} & \textbf{30.91} &\textbf{49.65} & \textbf{34.72} & \textbf{55.12}  \\

\cmidrule(lr){2-11}

& \multirow{4}{*}{BLIP-2} & Image-Only & \phantom{0}7.53 & 17.93 & \phantom{0}4.21 & 11.89 & 10.59 & 23.51 & \phantom{0}7.81 & 18.41 \\
& & Text Only & 24.68 & 43.59 & 20.77 & 41.64 & 25.95 & 42.83 & 27.33 & 46.31  \\
& & Image+Text & \underline{29.21} & \underline{50.05} & \underline{23.30} & \underline{45.61} & \underline{32.82} & \underline{53.09} & \underline{31.51} & \underline{51.45}  \\
& & SCOT (Ours) & \textbf{38.45} & \textbf{60.03} & \textbf{32.78} & \textbf{55.91} & \textbf{41.42} & \textbf{61.09} & \textbf{41.15} & \textbf{63.10} & \textbf{49.24} \\
\bottomrule
\end{tabular}
\vspace{0.3em}
\caption{\textbf{Results on FashionIQ.} 
Zero-shot results from our proposed approach compared against existing zero-shot methods (bottom) presented alongside some fully-supervised approaches (top). For fair comparisons, SEARLE results are from the inversion model and TransAgg results are using frozen backbones. See supplementary material for more results.
}
\label{table:fashioniq}
\vspace{-1.0em}
\end{table*}


\begin{table*}[t]
\centering
\footnotesize
\begin{tabular}{lllccccccc}
\toprule
\multirow{2}{*}{} & \multirow{2}{*}{\textbf{Method}} & \multirow{2}{*}{\textbf{Backbone}} & \multicolumn{4}{c}{\textbf{Recall@$K$}} & \multicolumn{3}{c}{\textbf{Recall$_{\text{subset}}$@$K$}} \\
\cmidrule(lr){4-7} \cmidrule(lr){8-10}
 & & & \( K = 1 \) & \( K = 5 \) & \( K = 10 \) & \( K = 50 \) & \( K = 1 \) & \( K = 2 \) & \( K = 3 \) \\
\midrule
\parbox[t]{15pt}{\multirow{4}{*}{\rotatebox[origin=c]{90}{Superv.}}}

& Multi & MAAF \cite{maaf} & 10.31 & 33.03 & 48.30 & 80.06 & 21.05 & 41.81 & 61.60 \\ 
& OSCAR & CIRPLANT \cite{CIRR} & 19.55 & 52.55 & 68.39 & 92.38 & 39.20 & 63.03 & 79.49 \\ 
& CLIP L/14 & CLIP4CIR \cite{combinerarch} & 33.59 & 65.35 & 77.35 & 95.21 & 62.39 & 81.81 & 92.02 \\
& BLIP & BLIP4CIR \cite{BLIP4Cir} & 40.15 & 73.08 & 83.88 & 96.27 & 72.10 & 88.27 & 95.93 \\
\midrule

\parbox[t]{15pt}{\multirow{23}{*}{\rotatebox[origin=c]{90}{Zero-Shot}}}

& \multirow{7}{*}{CLIP B/32} & Image-only & \phantom{0}6.94 & 22.94 & 33.71 & 59.18 & 21.06 & 41.01 & 60.34\\
& & Text-only & 21.16 & 45.35 & 57.40 & 81.06 & \textbf{62.26} & \textbf{81.08} & \textbf{90.75} \\
& & Image+Text & 10.46 & 32.41 & 46.39 & 75.11 & 30.09 & 54.24 & 73.20\\
& & PALAVRA \cite{PALAVRA} & 16.62 & 43.49 & 58.51 & 83.95 & 41.61 & 65.30 & 80.94 \\
& & SEARLE \cite{SEARLE} & \underline{24.00} & \underline{53.42} & \underline{66.82} & \underline{89.78} & 54.89 & 76.60 & 88.19\\
& & TransAgg \cite{transagg} & \textbf{24.46} & \textbf{53.61} & \textbf{67.54} & \textbf{89.81} & \underline{57.81} & \underline{78.17} & \underline{89.54} \\
& & SCOT (Ours) & 22.80 & 53.18 & 66.22 & 89.64 & 53.25 & 75.45 & 88.31 \\

\cmidrule(lr){2-10}

& \multirow{7}{*}{CLIP L/14} & Image-only & \phantom{0}7.47 & 23.88 & 34.07 & 57.57 & 20.87 & 41.95 & 61.13\\
& & Text-only & 22.00 & 45.79 & 57.57 & 79.59 & \textbf{61.71} & \textbf{80.26} & \textbf{90.43} \\
& & Image+Text & 10.55 & 32.70 & 45.71 & 74.26 & 31.06 & 55.69 & 73.93\\
& & Pic2Word \cite{Saito_2023_CVPR} & 23.9\phantom{0} & 51.7\phantom{0} & 65.3\phantom{0} & 87.8\phantom{0} & - & - & - \\
& & SEARLE-XL \cite{SEARLE} & 24.24 & 52.48 & 66.29 & 88.84 & 53.76 & 75.01 & 88.19 \\
& & TransAgg \cite{transagg} & \textbf{25.04} & \textbf{53.98} & \textbf{67.59} & \underline{88.94} & \underline{55.33} & \underline{76.82} & \underline{88.94}\\
& & SCOT (Ours) & \underline{24.36} & \underline{53.52} & \underline{67.37} & \textbf{89.35} & 51.47 & 74.24 & 87.90 \\

\cmidrule(lr){2-10}

& \multirow{5}{*}{BLIP} & Image-only & \phantom{0}7.23 & 25.78 & 37.35 & 62.34 & 20.60 & 40.96 & 61.35 \\
& & Text-only & 34.19 & 61.68 & 71.74 & 87.83 & \textbf{72.34} & \textbf{87.97} & \textbf{94.79} \\
& & Image+Text & \phantom{0}8.24 & 28.96 & 41.23 & 68.07 & 23.64 & 45.35 & 66.29 \\
& & TransAgg \cite{transagg} & \underline{34.89} & \underline{64.75} & \underline{76.24} & \underline{92.22} & \underline{66.34} & \underline{83.76} & \underline{92.92} \\
& & SCOT (Ours) & \textbf{36.31} & \textbf{66.19} & \textbf{77.37} & \textbf{92.96} & 64.73 & 83.20 & 92.15 \\

\cmidrule(lr){2-10}

& \multirow{4}{*}{BLIP-2} & Image-only & \phantom{0}7.59 & 24.43 & 35.56 & 61.42 & 20.74 & 40.67 & 61.08 \\
& & Text-only & \underline{33.52} & \underline{61.50} & \underline{71.35} & \underline{88.31} & \underline{72.53} & \underline{88.02} & \textbf{94.87} \\
& & Image+Text & 19.69 & 49.98 & 64.39 & 90.01 & 45.69 & 71.18 & 85.83 \\
& & SCOT (Ours) & \textbf{36.82} & \textbf{64.34} & \textbf{74.48} & \textbf{93.42} & \textbf{75.73} & \textbf{88.70} & \underline{94.84} \\
\bottomrule
\end{tabular}
\vspace{0.2em}
\caption{
    \textbf{Results on CIRR.}
    Zero-shot results from our proposed approach compared against existing zero-shot methods (bottom) presented alongside some fully-supervised approaches (top). 
For fair comparisons, SEARLE results are from the inversion model and TransAgg results are using frozen backbones. See supplementary material for more results.
}
\label{table:cirr}
\vspace{-1.5em}
\end{table*}    

\subsection{Comparison with state-of-the-art methods}
\noindent \textbf{Evaluation metrics.} We present a quantitative comparison against the state-of-the-art on the FashionIQ~\cite{fashionIQ} and CIRR~\cite{CIRR} datasets. The evaluation metric for FashionIQ is the average recall at rank $K$ (R@$K$). Following prior work~\cite{SEARLE, Saito_2023_CVPR, transagg} we present R@10 and R@50 on the validation set.
For CIRR, we follow the authors' proposed protocol to report Recall@$K$ at four different ranks, i.e., $K \in \{1, 5, 10, 50\}$, along with Recall$_{\text{subset}}$@$K$, which uses small subsets with fully labelled negatives for each query image~\cite{CIRR}. We show results for existing zero-shot approaches and fully-supervised approaches.

\noindent \textbf{Baselines.} As reference, we present results of retrieving using just the image embedding (Image-Only), just the modification text embedding (Text-Only), or the sum of the two (Image+Text). For a fair comparison against prior zero-shot methods such as Pic2Word~\cite{Saito_2023_CVPR} and SEARLE~\cite{SEARLE}, which rely on frozen backbones, we include results from TransAgg~\cite{transagg} with frozen backbones. Baldrati et al.~\cite{SEARLE} present two variants of their approach: SEARLE-OTI, which requires inference-time optimization, and SEARLE, which trains a textual inversion network to reproduce the OTI outputs in a single forward pass. Here, we use the reported results for SEARLE and its larger version SEARLE-XL. Finally, we note that existing methods use different backbones, amounts and types of data, fusion architectures, and pretraining strategies. For instance, Pic2Word \cite{Saito_2023_CVPR} uses 3M images with a frozen CLIP L/14 backbone within a textual inversion-based approach, whereas TransAgg \cite{transagg} uses 32K synthetic triplets with BLIP and a Transformer-based fusion method. We provide results segregated by backbone in Tables \ref{table:fashioniq} and \ref{table:cirr}, and further analyze the importance of different contrastively-trained backbones in Section \ref{sec:backbones}.


\noindent \textbf{Results on FashionIQ.} From Table~\ref{table:fashioniq}, the best-performing SCOT model improves by 11.78\% on R@10 and by 13.8\% on R@20 over SEARLE-XL\cite{SEARLE}. SCOT also demonstrates notable data efficiency: utilizing only 290K image-text pairs for training, in contrast to the 3M images used in Pic2Word's training with Conceptual Captions, we achieve 13.75\% improvement over Pic2Word on R@10.
The zero-shot performance of SCOT exceeds many fully-supervised methods, such as DCNet~\cite{dcnet}, CLIP4Cir~\cite{combinerarch}, and FashionVLP~\cite{fashionVLP}, while approaching that of BLIP4CIR~\cite{BLIP4Cir}. 

\noindent \textbf{Results on CIRR.} From Table~\ref{table:cirr}, SCOT exhibits improvements of 
12.58\% at R@1 and 10.86\% at R@5 over SEARLE-XL. 
We also see that Text-Only performance is significantly higher than Image-Only performance on CIRR, and that naively adding image features to text degrades performance. This is explained by a known shortcoming of CIRR ---also noted in prior work~\cite{SEARLE}---that modification texts often describe the target image completely, with reference images providing no additional information. 


\begin{figure}
\centering
\includegraphics[scale=0.20]{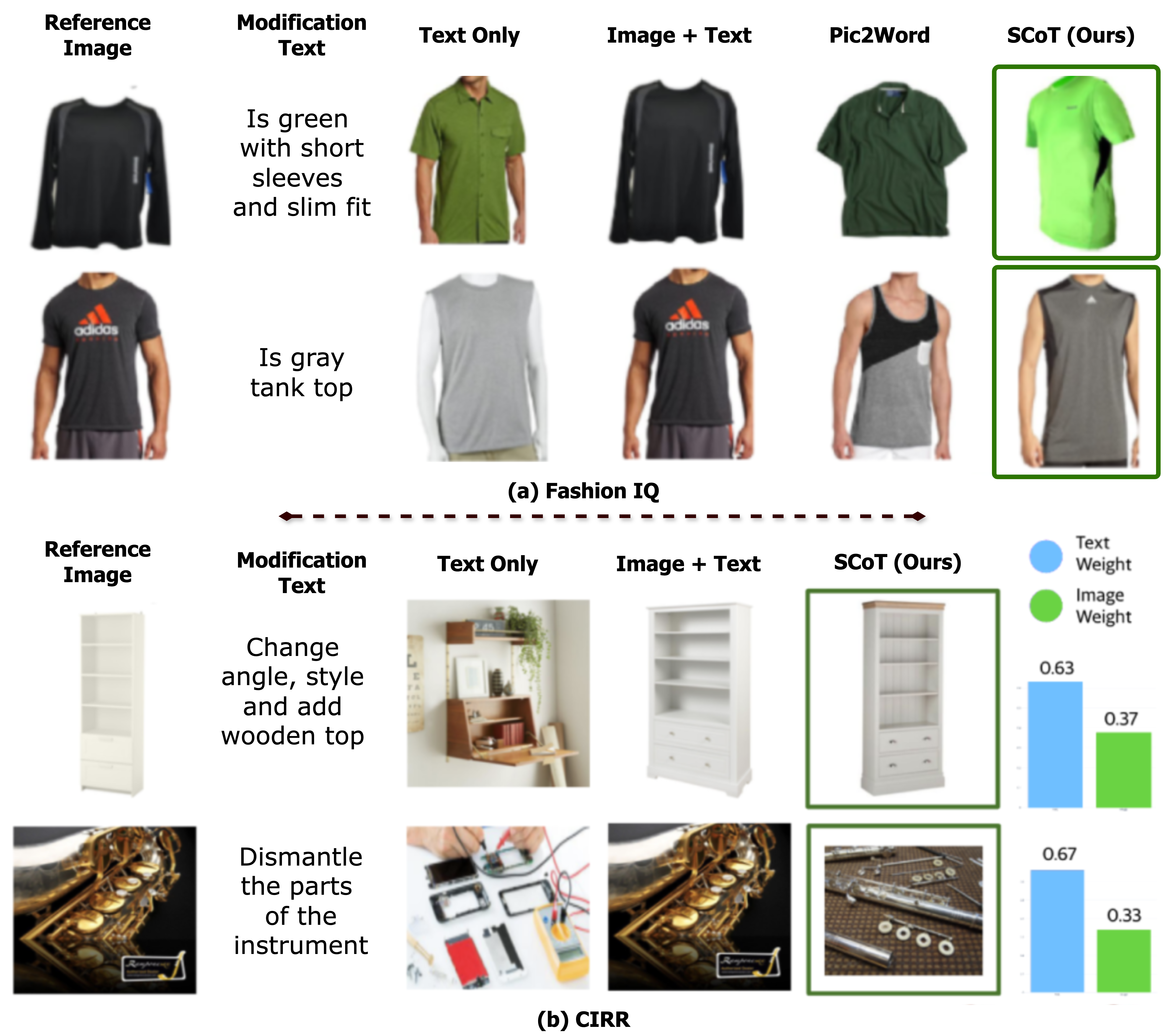}
\caption{\textbf{Qualitative retrieval results} on validation sets. \textbf{Top:} FashionIQ \cite{fashionIQ}. \textbf{Bottom:} CIRR \cite{CIRR}. A green box indicates the correctly retrieved image. For CIRR, the rightmost column illustrates the corresponding modality weight learned by SCOT for that example. (Best viewed in color.)}
\label{fig:fiq_qualitative}
\vspace{-2.0em}
\end{figure}



\begin{figure}
\centering
\begin{subfigure}[b]{0.36\textwidth}
\includegraphics[width=\textwidth]{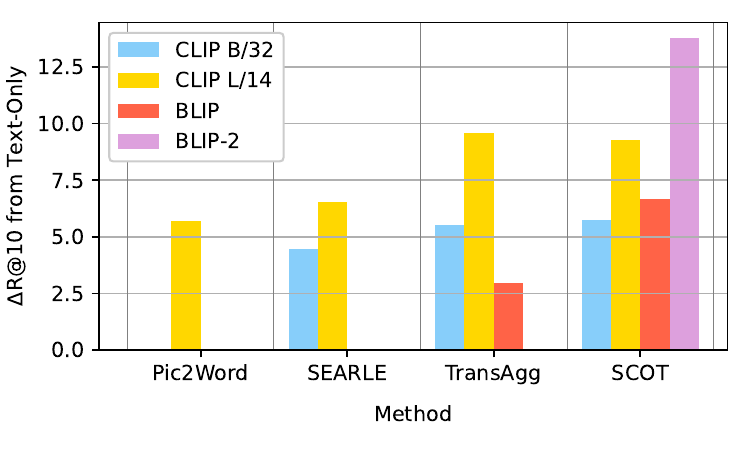}
\vspace{-1.0em}
\caption{FashionIQ: $\mathrm{\Delta}$R@10 from Text-Only}
\label{fiq_deltas}
\end{subfigure}
\begin{subfigure}[b]{0.36\textwidth}
\includegraphics[width=\textwidth] {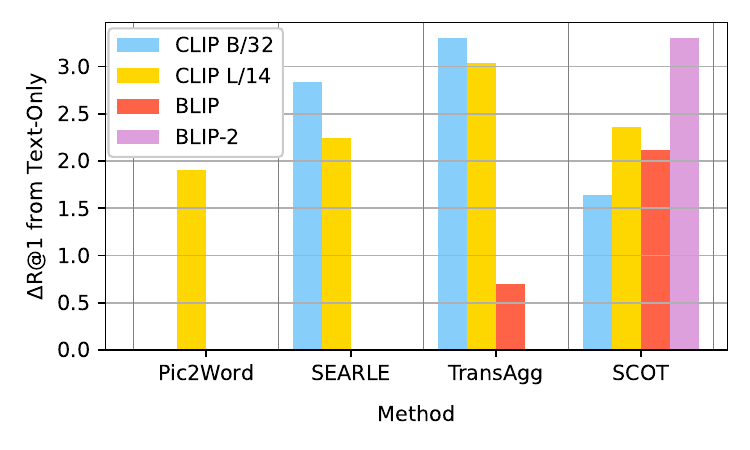}
\vspace{-1.0em}
\caption{CIRR: $\mathrm{\Delta}$R@1 from Text-Only}
\label{cirr_deltas}
\end{subfigure}
\vspace{-0.5em}
\caption{\textbf{Gains relative to Text-Only.} Difference in recall ($\mathrm{\Delta}$R) between methods and the backbone-matched Text-Only baseline.}
\label{fig:backbone_scaling_curves}
\vspace{-0.3em}
\end{figure}

\subsection{Discussion}
\label{sec:discussion}


\noindent \textbf{1. Qualitative analysis.} In Fig.~\ref{fig:fiq_qualitative} (Top) we present zero-shot qualitative retrieval results on FashionIQ, illustrating domain-specific behavior. The figure shows that SCOT effectively composes images and text to retrieve the most accurate product image. The second row is particularly interesting: all methods retrieve a gray tank top, but only SCOT specifically retrieves one with the Adidas logo, which was also present in the reference image. We evaluate the qualitative performance on open-world images using CIRR in Fig.~\ref{fig:fiq_qualitative} (Bottom). As discussed earlier, often in CIRR the modification text can be informative enough to retrieve the correct target image. 
The learned dynamic scalar scores of the Combiner network are shown in the last column. In cases where the modification text completely describes the target image---such as in the third row---SCOT assigns a high weight to the text representation. In last row, it can be observed that the dog breed can only be inferred through the reference image; consequently SCOT assigns nearly equal weight to both image and text representations.

\noindent \label{sec:backbones}\textbf{2. Impact of image-text alignment backbones.}
Using text embeddings as a proxy for image embeddings requires the image and text embedding spaces to be well-aligned. 
Here, we study the behavior of SCOT and other methods as we vary the encoder backbones.
To recall, based on previous results~\cite{CLIPopenai, li2022blip1, li2023blip2}, the relative ranking of the backbones we experimented with is CLIP-B/32 $<$ CLIP-L/14 $<$ BLIP $<$ BLIP-2.
From Table~\ref{table:fashioniq}, with CLIP B/32, SCOT gets an average R@10 of 24.14\% on FashionIQ \cite{fashionIQ}. With CLIP L/14, we observe 28.27\%, nearly 4\% higher. With BLIP, we observe another 2\% improvement at R@10, while TransAgg produces a {\it drop} of 1.6\%. Finally, for SCOT with BLIP-2, we see the largest improvement, of 8\% over BLIP. 
On CIRR, as seen in Table~\ref{table:cirr}, when using the CLIP B/32 backbone, SCOT is behind both TransAgg and SEARLE. SCOT then surpasses SEARLE when using the CLIP L/14 backbone, and surpasses TransAgg when switching to the BLIP backbone. Thus, as with FashionIQ, the relative performance of methods changes with different backbones, with SCOT's advantage increasing as the backbones improve.
%
This can be more clearly seen in Fig.~\ref{fig:backbone_scaling_curves} where we present the relative gains of different methods with respect to the `Text-Only' baseline. 
We define relative gain $\mathrm{\Delta}$R@$K$ as the difference between Recall@$K$ of a given method and that of the `Text-Only' baseline with the corresponding backbone. 
On both FashionIQ and CIRR,
Fig.~\ref{fig:backbone_scaling_curves} shows that as we improve the backbones, the relative gain of SCOT increases. Thus, not only does SCOT benefit from better backbones as represented by the performance of the `Text-Only' baseline on those backbones, but its gain over that baseline also increases.
Of note, with the BLIP backbone, SCOT has relative gains that are 2-3 larger than that of TransAgg with BLIP, showing that SCOT is unique in obtaining higher relative gains with better backbones.
  
\begin{figure}[t]
\centering
\includegraphics[scale=0.11]{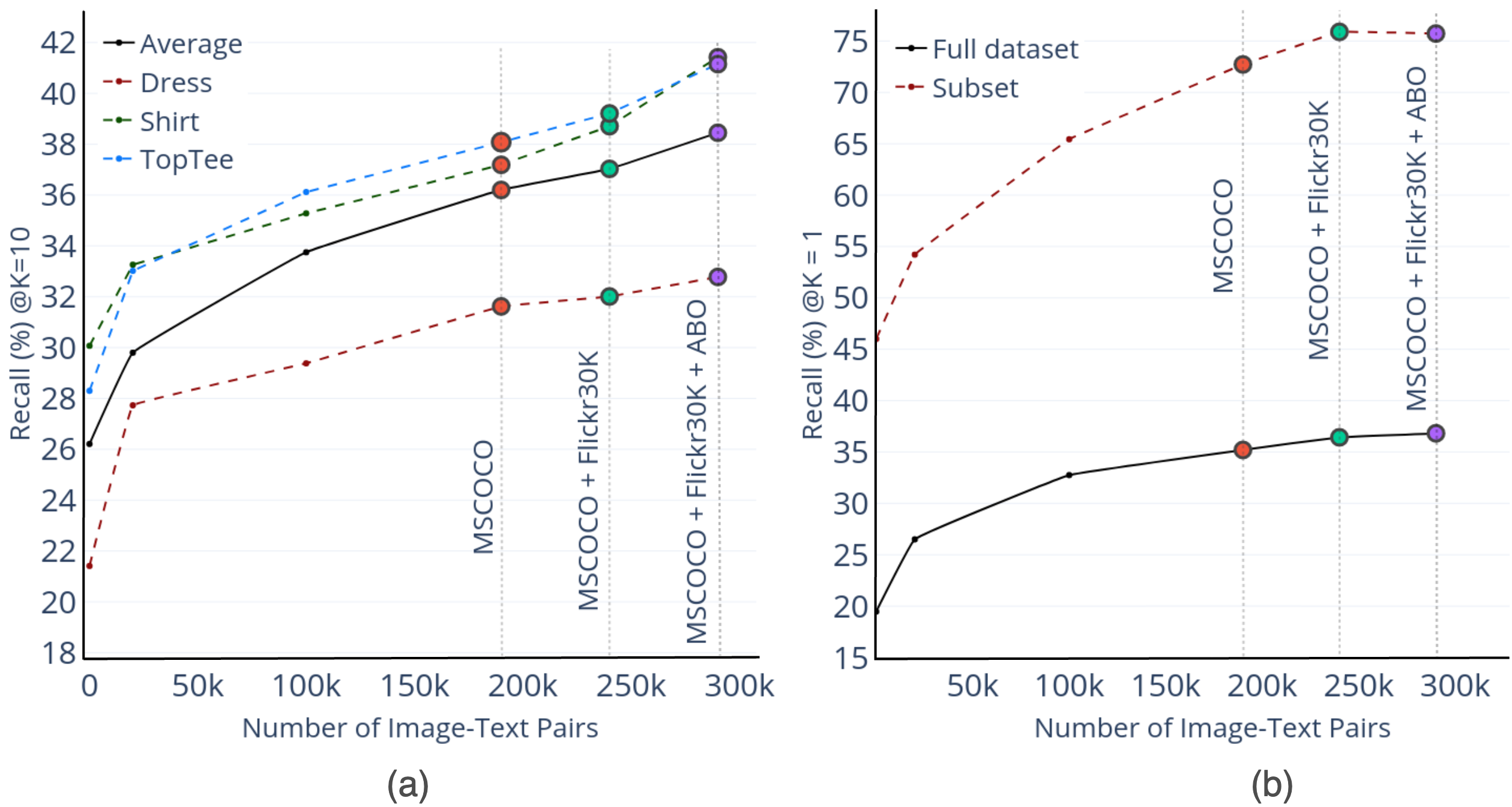}
\vspace{-0.5em}
\caption{\textbf{Performance when changing the size and distribution of the training set}, evaluated on the (a) FashionIQ~\cite{fashionIQ} validation set across clothing types and (b) CIRR~\cite{CIRR} test set.
}
\label{fig:scalingcurve}
\vspace{-1.3em}
\end{figure}

\noindent \textbf{3. Impact of dataset distribution.} In Fig.~\ref{fig:scalingcurve}, we illustrate how performance changes as we expand the training set. On both FashionIQ and CIRR, recall increases when utilizing larger subsets of the 189K MSCOCO image-caption pairs. This trend continues with the addition of Flickr30K. While both MSCOCO and Flickr30K contain generic real-world images, we wanted to also evaluate improvements brought by including domain-specific images. The Amazon Berkeley Objects (ABO)~\cite{ABO} dataset contains a variety of retail products, such as phone cases and furniture, accompanied by detailed captions. By including 58K image-caption pairs from ABO, we see around a 1\% improvement on FashionIQ's average R@10, going from 37.21\% to 38.45\%. 
Specifically for Shirt and Top/Tee, performance improves by around 2\% when adding ABO. On CIRR, as shown in Fig.~\ref{fig:scalingcurve}(b), incorporating ABO yields only a marginal gain in R@1 and no improvement in R$_{\text{subset}}$@$1$, likely due to the differing image distributions between ABO and CIRR.

\begin{figure}[t]
\centering
\includegraphics[scale=0.50]{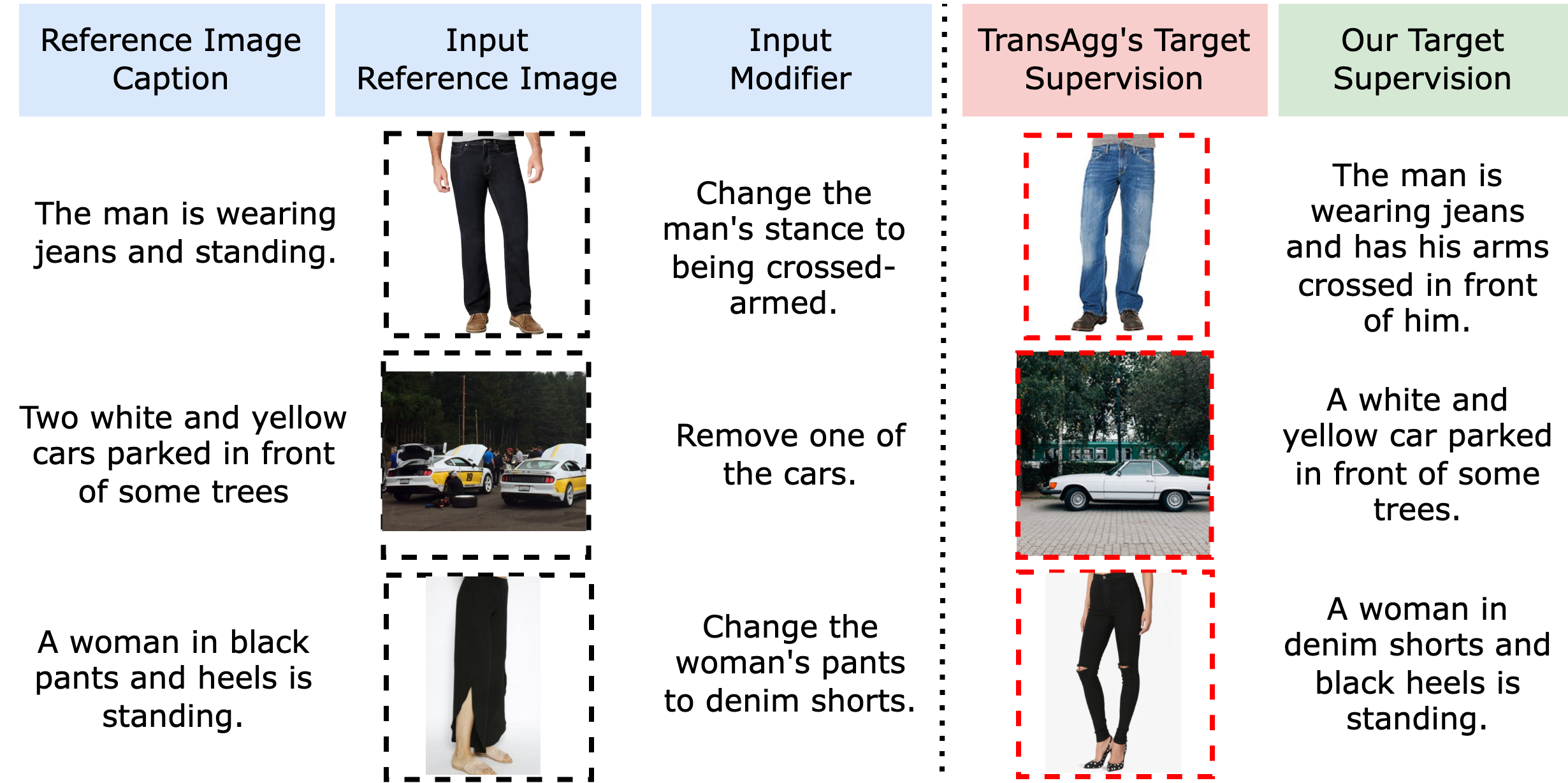}
\caption{\textbf{Examples from LAION-CIR-LLM~\cite{transagg}} illustrating the challenges of using retrieved images as target supervision.}
\label{fig:transagg_images}
\vspace{-1.2em}
\end{figure}

\begin{table}[t]
\centering
\setlength{\tabcolsep}{2.7pt}
\footnotesize

\begin{tabular}{cccccccccc}
\toprule
\multirow{2}{*}{\textbf{Supervision}} & \multicolumn{2}{c}{\textbf{Average}} & \multicolumn{2}{c}{\textbf{Dress}} & \multicolumn{2}{c}{\textbf{Shirt}} & \multicolumn{2}{c}{\textbf{Top/Tee}} \\
\cmidrule(lr){2-3} \cmidrule(lr){4-5} \cmidrule(lr){6-7} \cmidrule(lr){8-9}
& \textbf{\scriptsize R@10} & \textbf{\scriptsize R@50} & \textbf{\scriptsize R@10} & \textbf{\scriptsize R@50} & \textbf{\scriptsize R@10} & \textbf{\scriptsize R@50} & \textbf{\scriptsize R@10} & \textbf{\scriptsize R@50} \\
\midrule
Image \cite{transagg} & 29.02 & 50.49 & 22.65 & 45.06 & 33.21 & 53.28 & 31.20 & 53.13 \\
Text (Ours) & \textbf{35.17} & \textbf{56.16} & \textbf{29.54} & \textbf{50.96} & \textbf{36.45} & \textbf{57.26} & \textbf{39.52} & \textbf{60.27} \\
\bottomrule
\end{tabular}
\vspace{0.3em}
\caption{\textbf{FashionIQ results with different supervision targets} when training on LAION-CIR-LLM~\cite{transagg} with a BLIP-2 backbone.
We observe that the use of text targets for supervision performs significantly better than the image targets available in the dataset.}
\label{table:transagg_dataset}
\vspace{-1.5em}
\end{table}

\noindent \textbf{4. Text supervision vs retrieved image supervision.} 
An alternative way of using LLM-generated text triplets for ZS-CIR involves using each of the generated modified captions as a query to retrieve an image from a large corpus. Each retrieved image is then used as target supervision for its corresponding reference image and generated modification text. Concurrently to our work, Liu et al.~\cite{transagg} experimented with this type of approach.
Fig.~\ref{fig:transagg_images} displays examples from the LAION-CIR-LLM dataset they proposed, which is based on image-caption pairs taken from LAION-COCO~\cite{laioncoco}.\footnote{LAION-COCO (and by extension LAION-CIR-LLM) contains many clothing and product images, resulting in good coverage over FashionIQ.} As shown in the figure, the retrieved target images often do not match the expected modified caption due the absence of a relevant image in the corpus and/or retrieval errors.
Table~\ref{table:transagg_dataset} presents an experiment comparing the use of the retrieved image target supervision from LAION-CIR-LLM~\cite{transagg} against text supervision using the dataset's modified captions. The experiment uses BLIP-2~\cite{li2023blip2} as image and text encoder, and the Combiner~\cite{combinerarch} as composition function.
We see that using retrieved images as targets gives an average R@10 on FashionIQ of 29.02\%, whereas using text targets as proposed in our approach achieves 35.17\%.

\section{Conclusion}
We propose a novel approach towards annotation-free ZS-CIR which leverages existing large captioned image datasets, along with contrastively-pretrained vision-language models. We demonstrate the zero-shot generalizability of this technique through extensive experimentation on domain-specific and open-world datasets. Our proposed approach, SCOT, achieves state-of-the-art performance in zero-shot settings while being on par with various fully-supervised approaches. We further substantiate this work with qualitative and quantitative experiments to analyze the impact of various components of our pretraining strategy.

\bibliographystyle{splncs04}
\bibliography{main}

\clearpage

\end{document}